\documentclass[letterpaper]{article} 
\usepackage{aaai25}  
\usepackage{times}  
\usepackage{helvet}  
\usepackage{courier}  
\usepackage[hyphens]{url}  
\usepackage{graphicx} 
\usepackage{natbib}  
\usepackage{caption} 
\usepackage{algorithm}
\usepackage{algorithmic}

\usepackage{newfloat}
\usepackage{listings}
\DeclareCaptionStyle{ruled}{labelfont=normalfont,labelsep=colon,strut=off} 
\floatstyle{ruled}
\newfloat{listing}{tb}{lst}{}
\floatname{listing}{Listing}
\usepackage{graphicx} 
\usepackage{natbib}  
\usepackage{caption} 
\usepackage{babel}
\usepackage{algorithm}
\usepackage{algorithmic}
\usepackage[most]{tcolorbox}
\AtBeginEnvironment{tcolorbox}{\small}
\usepackage{newfloat}
\usepackage{listings}
\DeclareCaptionStyle{ruled}{labelfont=normalfont,labelsep=colon,strut=off} 
\floatstyle{ruled}
\newfloat{listing}{tb}{lst}{}
\floatname{listing}{Listing}
\usepackage{subcaption}

\title{
ElectroVizQA: How well do Multi-modal LLMs perform in Electronics Visual Question Answering?
}
\author {
    Pragati Shuddhodhan Meshram, 
    Swetha Karthikeyan, 
    Bhavya, 
    Suma Bhat 
}
\affiliations {
    University of Illinois Urbana-Champaign\\
    \{ psm12, swethak2, bhavya2, spbhat2 \} @illinois.edu
}

\begin{document}

\maketitle

\begin{abstract}

Multi-modal Large Language Models (MLLMs) are gaining significant attention for their ability to process multi-modal data, providing enhanced contextual understanding of complex problems. MLLMs have demonstrated exceptional capabilities in tasks such as Visual Question Answering (VQA); however, they often struggle with fundamental engineering problems, and there is a scarcity of specialized datasets for training on topics like digital electronics. To address this gap, we propose a benchmark dataset called ElectroVizQA specifically designed to evaluate MLLMs' performance on digital electronic circuit problems commonly found in undergraduate curricula. This dataset, the first of its kind tailored for the VQA task in digital electronics, comprises approximately 626 visual questions, offering a comprehensive overview of digital electronics topics. This paper rigorously assesses the extent to which MLLMs can understand and solve digital electronic circuit questions, providing insights into their capabilities and limitations within this specialized domain. By introducing this benchmark dataset, we aim to motivate further research and development in the application of MLLMs to engineering education, ultimately bridging the performance gap and enhancing the efficacy of these models in technical fields. Our data and code are available here.\footnote{\url{https://github.com/Pragati-Meshram/ElectroVizQA}}
\end{abstract}
\section{Introduction} The recent shift from single-modal models to multi-modal models aims to leverage diverse information sources by incorporating different modalities. This transition has brought about remarkable advancements in models like GPT-4o with demonstrated improvements compared to its predecessors on standard reasoning and related STEM benchmarks such as MATH \cite{hendrycks2021measuring}, GSM-8K \cite{cobbe2021gsm8k}, ScienceQA \cite{lu2022scienceqa}, MMLU \cite{hendrycks2021mmlu}.\\
As these models continue to evolve, their application to Visual Question Answering (VQA) has gained considerable attention. Extensive research has focused on VQA, where multimodal large language models (MLLM) such as InstructBlip \cite{dai2023instructblip}, Llava \cite{liu2023llava}, Sphinx \cite{lin2023sphinx}, and GPT4-o \cite{gpt4o} exhibit strong performance across various tasks in diverse domains. 
Recently released benchmarks like MathVista \cite{lu2023mathvista} and systematic studies like MathVerse \cite{zhang2024mathverse} have provided comprehensive evaluations of MLLMs over VQA tasks. Beyond these, domain-specific datasets, such as those in Llava-Med \cite{liu2023llava-med}, JEE-Bench \cite{arora2023have},  and IconVQA \cite{li2023iconvqa}  further illustrate the breadth of data utilized for fine-tuning these models to set challenging benchmarks.\\
Building on these advancements in multi-modal learning, we focus on digital electronics, a foundational subject in engineering education and electronic design automation, where problem solving relies on a strong reasoning ability. For instance, the logical operations performed by digital gates AND, OR, NOT, XOR, XNOR have direct mathematical analogies, underscoring the fundamental nature of this subject. Beyond the reasoning aspect, answering these questions requires a systematic collation of information from tables, associated figures and text, thus posing definite challenges for multi-modal models. 
As  students in all domains increasingly rely on MLLMs like ChatGPT, it is crucial for these models to provide reliable solutions in this fundamental field. This paper presents a manually created and curated dataset tailored to fill a benchmarking gap and evaluates MLLMs' performance on answering these problems about digital electronic circuits. 




\begin{figure}[t]
\centering
\includegraphics[trim=10 12 10 10,clip,width=0.9\columnwidth]{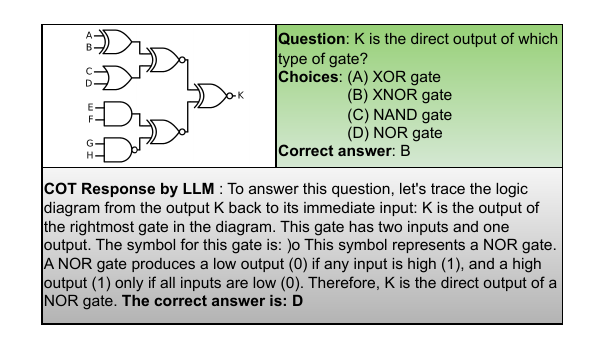} 
\caption{Example illustrating the challenges faced by LLMs in recognizing basic logic gates accurately
}
\label{example}
\end{figure}
The \textbf{ElectroVizQA} dataset is developed along three problem dimensions. (1) The \textit{Conceptual dimension} captures fundamental concepts for solving digital electronics problems, such as Karnaugh Map (K-map), Truth Table; (2) The \textit{Visual context dimension} relates to the visual elements in the dataset and spans the topics such as finite state machines (FSM), gates, encoders/decoders; and finally the \textit{Solving strategy dimension} relates to the strategy required to solve the problem---factual, computational, or deep analysis requiring collating knowledge and deeper reasoning. Our dataset incorporates fine-grained question category labels derived from these problem dimensions, meticulously integrating textual and visual elements to assess multi-modal large language models in both visual and textual understanding within question-answering tasks. In addition to constructing the dataset, we benchmark LLMs and provide rigorous error analysis of their outputs.

To develop this benchmark, we referenced two resources to curate the questions with figures. The first is a set of course notes\footnote{\url{http://lumetta.web.engr.illinois.edu/120-S17/ece120-spring-2017-notes-for-students.pdf }\label{t1}} used in an introductory course on fundamental digital electronics for undergraduate students at a large U.S. public university and referenced with the author's permission. The second textbook\footnote{\url{https://textbookequity.org/Textbooks/TBQ_Feher_DigitalLogic.pdf (Creative Commons Attribution 3.0 License)} \label{t2}} covers additional topics in the domain and has a Creative Commons attribution license. Together, these textbooks provide a comprehensive coverage of the essential digital electronics topics, including foundational elements such as Karnaugh maps, truth tables, and combinational logic circuits spanning decoders, multiplexers, latches, flip-flops, counters, and finite state machines. Additionally, we utilize  schemdraw\footnote{\url{https://schemdraw.readthedocs.io/en/latest/index.html} \label{schemdraw}}, an online library, to draw circuits when the visual elements were needed.  We followed a systematic manual problem generation and review process to guarantee data quality. This resulted in 626 categorized questions from an initial pool of 800. 

This is followed by a quantitative and qualitative analysis of the outputs of state-of-the-art proprietary and open-source MLLMs on our dataset. Further, we designed prompts to identify error types and perform a critical analysis of LLMs' performance on our dataset.  Our findings reveal significant qualitative deficiencies, particularly in GPT4-o's understanding of visual content, despite its reasonable performance with textual information. 
In sum, our main contributions include the following: 
\begin{itemize}
\item We propose the first benchmark for digital elecronics VQA, ElectroVizQA, which has 626 meticulously created questions with manual annotations for three primary problem dimensions. We expect this benchmark to serve as a strong and reliable test bed and to foster future research on problem-solving with MLLMs. 
\item We conduct an extensive comparative analysis of MLLMs on our benchmark and investigate their visual and textual problem-solving capabilities.
\item To understand the challenges offered by our dataset, we conduct a careful manual error analysis of the MLLMs' responses to our VQA dataset.
\end{itemize}

\section{Related Work}

Recently, several benchmark datasets have been developed to test the STEM reasoning and problem-solving capabilities of large language models \cite{chang2024survey} mainly in the math and general science domains. GSM8k \cite{cobbe2021gsm8k} consists of grade school math word problems that require several steps of elementary arithmetic computations to solve. MATH dataset 
 \cite{hendrycks2021measuring} consists of challenging competition mathematics problems. MMLU \cite{hendrycks2021mmlu} covers 57 QA tasks including STEM domains like college mathematics, computer science, and physics. TheoremQA \cite{chen2023theoremqa} requires the application of mathematical theorems to solve questions.

Further,  a few benchmark datasets with visual questions for testing the multi-modal capabilities of LLMs in STEM problem-solving settings are also available. GeoQA \cite{chen2021geoqa} contains geometric problems in Chinese middle school exams along with visual diagrams. MathVista \cite{lu2023mathvista} includes mathematical reasoning on diagrams, logical reasoning on puzzles, statistical reasoning on functional plots, and scientific reasoning on academic figures. MathVerse \cite{zhang2024mathverse} focuses on only math visual problem solving and also provides step-by-step explanations for questions. ScienceQA \cite{lu2022learn} consists of elementary and high school multi-modal science questions. SciBench \cite{wang2023scibench} consists of more advanced, college-level scientific problems from mathematics, chemistry, and physics domains. 
\begin{figure*}[t]
\centering
\includegraphics[trim=10 20 10 10,clip, width=0.9\linewidth]{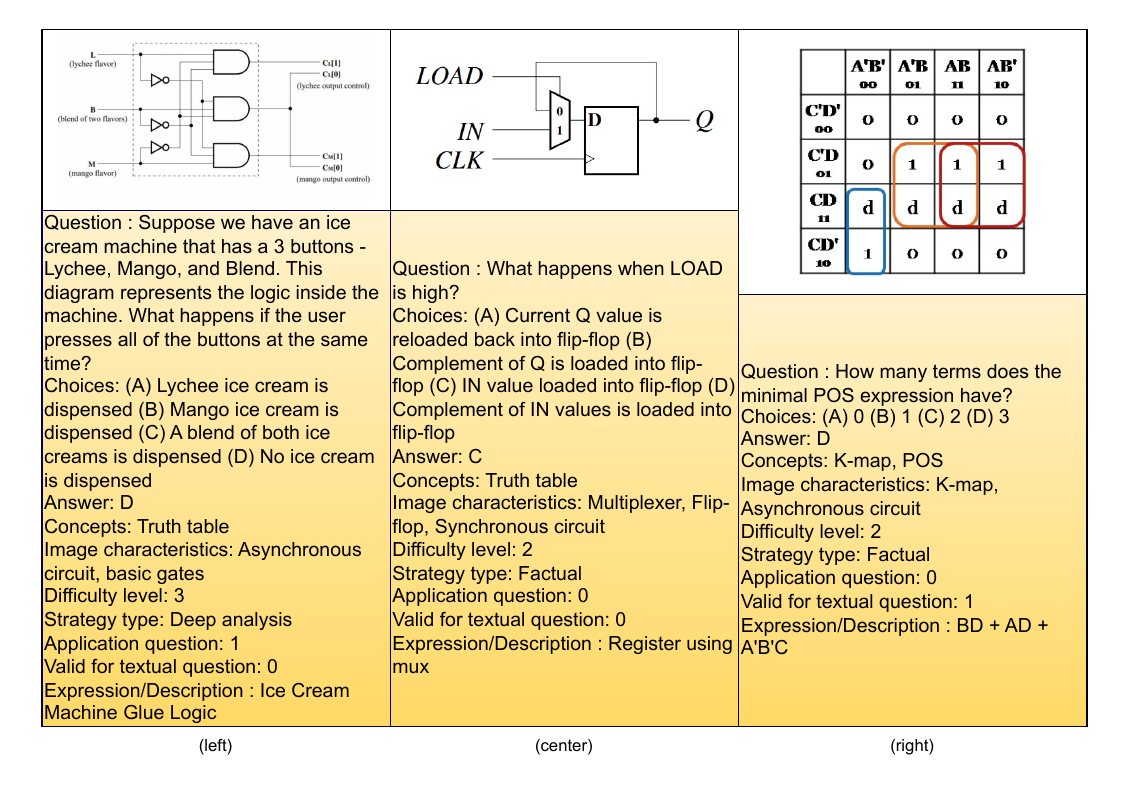} 
\caption{Examples of our annotated data
}
\label{data_example}
\end{figure*}
With the advancing capabilities of LLMs, more challenging benchmarks with engineering questions have recently been developed that contain some electronics questions. JEEBench \cite{arora2023have} contains pre-engineering questions from the IIT-JEE  exam. C-Eval \cite{huang2024c} presents a Chinese evaluation suite of questions across middle school, high school, college and professional grade levels in 52 diverse disciplines including electrical engineering. Both these datasets do not contain diagrams. A preliminary exploration of ChatGPT has also been done on solving four electrical circuit questions \cite{ogunfunmi2024exploration}. Closest to our work, MMMU \cite{yue2024mmmu} includes multi-modal, college-level questions spanning 30 subjects including electronics. However, the majority of their 291 electronics questions cover other topics, like analog electronics and electrical circuits, and only a handful of them are about digital electronics. To the best of our knowledge, ours is the first attempt to purposely create and use a dataset to perform an in-depth benchmarking study for this domain, which also includes fine-grained question category labels based on our proposed primary problem dimensions and a careful choice of the textual and visual components in the questions. 

\section{The ElectroVizQA dataset}\label{dataset}
\begin{figure*}
  \begin{minipage}[t]{.32\linewidth}
    \includegraphics[width=\linewidth]{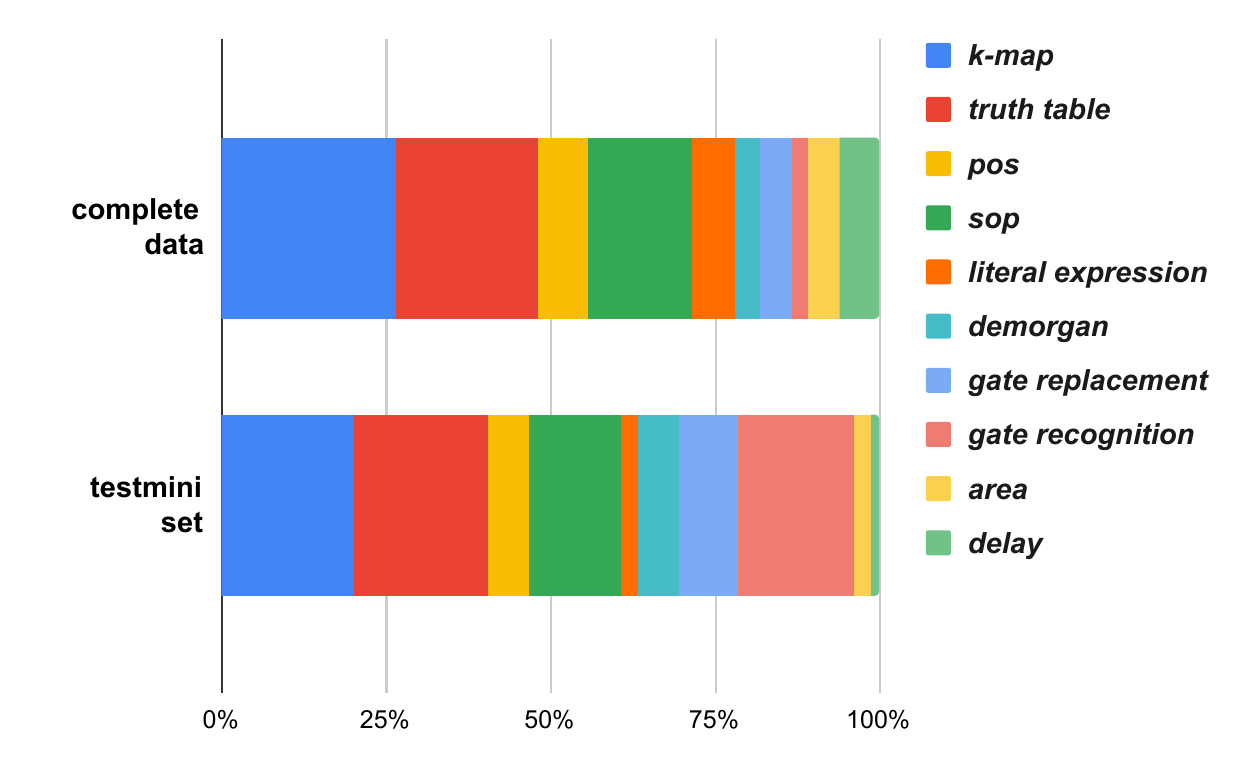}
    \subcaption{Basic concept}
    \label{fig4}
  \end{minipage}\hfil
  \begin{minipage}[t]{.32\linewidth}
    \includegraphics[width=\linewidth]{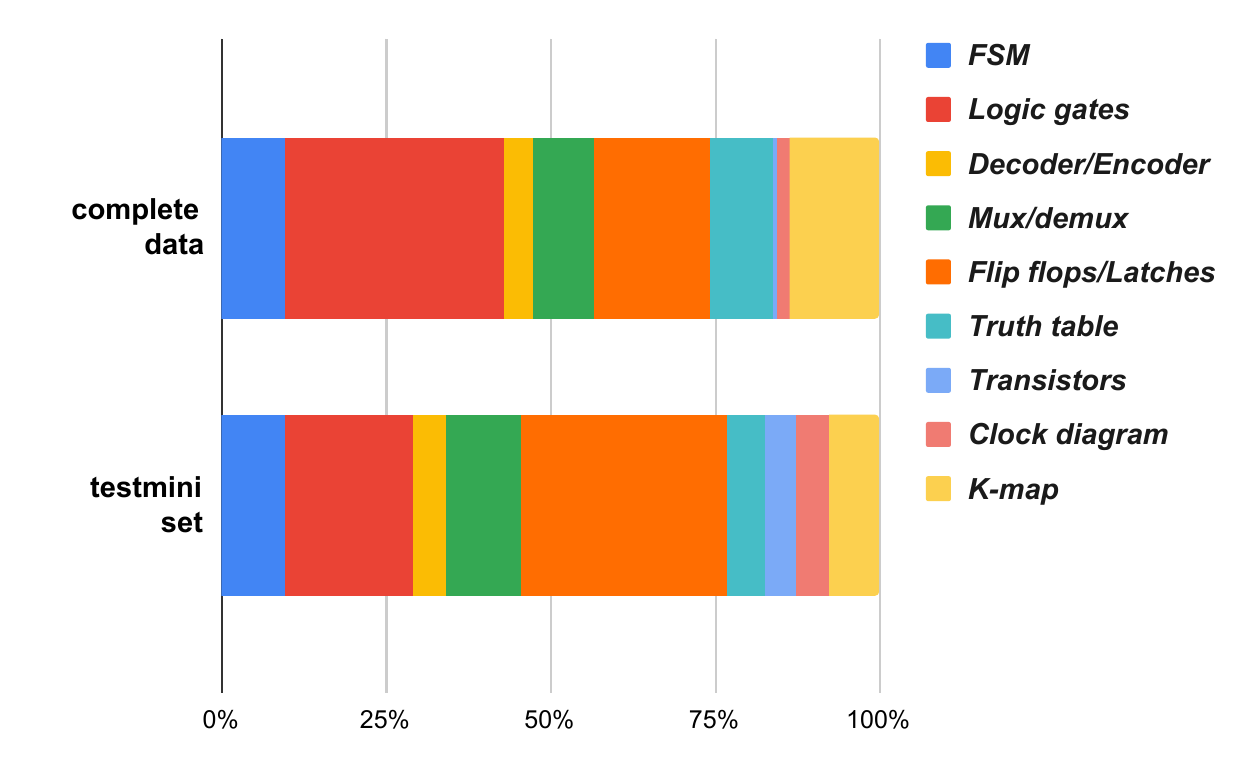}
    \subcaption{Visual context}
    \label{fig5}
  \end{minipage}\hfil
  \begin{minipage}[t]{.32\linewidth}
\includegraphics[width=\linewidth]{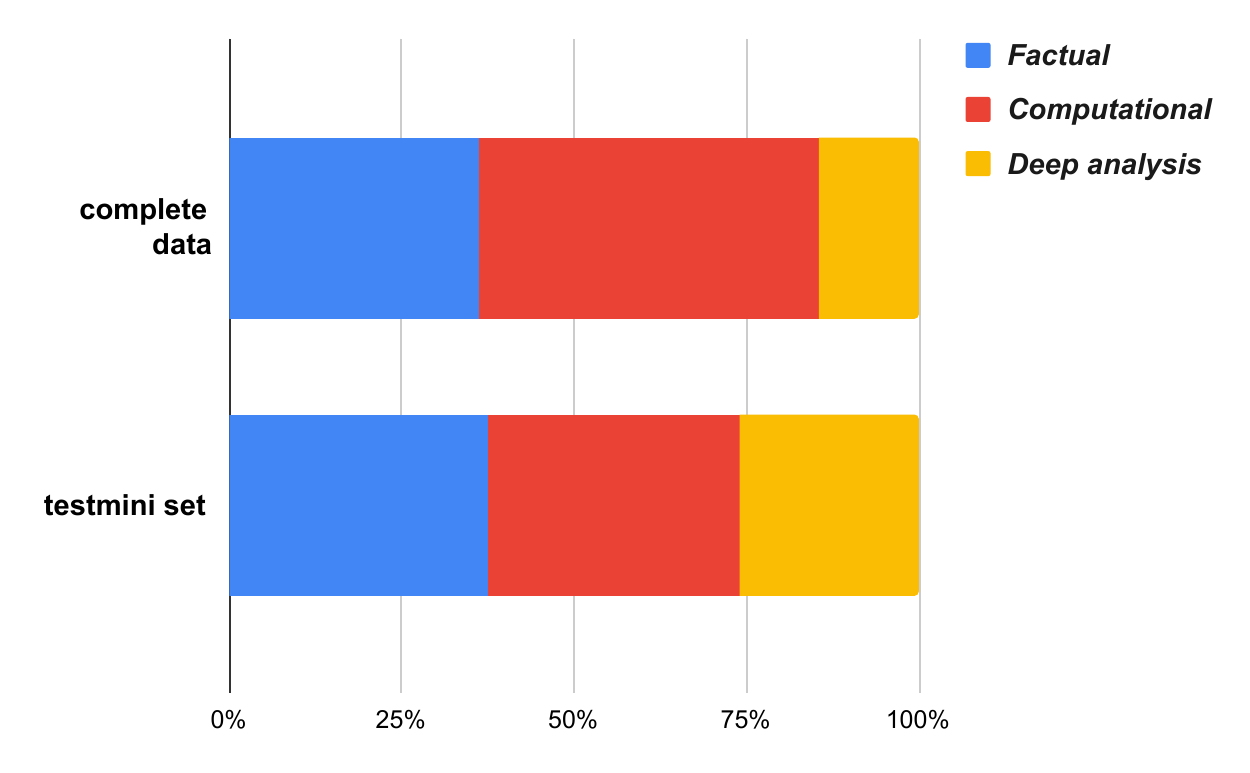}
\subcaption{Solving strategy}
\label{fig6}
\end{minipage}

\caption{(a), (b) and (c) represents the distribution of the three primary dimensions in the testmini set and the complete dataset}
\label{data_distribution}
\end{figure*}
\begin{figure}[h!]
\centering
\includegraphics[trim = 10 20 10 20 , width=0.9\columnwidth]{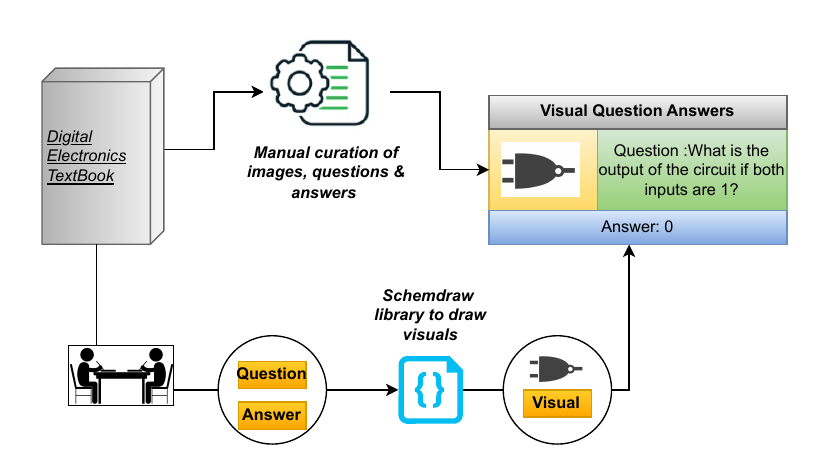} 
\caption{Pipeline for data creation}
\label{pipeline}
\end{figure}
Our dataset comprises 626 single-correct, multiple-choice \textbf{Electro}nics \textbf{Vis}ion \textbf{Q}uestion-\textbf{A}nswers, meticulously curated from the textbooks \cite{lumetta2017ece120, feher}

\noindent \textbf{Question Characteristics:}
Each question is annotated with both single-valued and multi-valued labels, which together constitute the metadata for each question.

\begin{itemize}
    \item \textbf{Multi-valued labels}: These are the represented \textit{Concepts} and \textit{Image Characteristics}, permitting multiple values per question, thereby capturing the complex nature of the problems. 
    \item \textbf{Single-valued labels}: These include \textit{Solving Strategy},
    \textit{Application Question}, \textit{Valid for Text Only}, \textit{Expression/Description}, and \textit{Difficulty Level}. The \textit{Expression/Description} field provides a minimalistic verbal representation of the visuals, while the \textit{Valid for Text Only} field designates whether a question can be satisfactorily answered using only textual question and \textit{Expression/Description}. These labels facilitate problem choice for language-only models' evaluation. An example of this is in Figure \ref{data_example} (right). Although we provide difficulty levels that  are based on annotator judgment, we don't include them in our analyses.
\end{itemize}
Our dataset is constructed such that the textual and visual components are mutually exclusive, ensuring that textual information does not elucidate the content of the visuals, thus challenging LLMs to independently extract and interpret visual information.


Further, to eliminate answer choice bias, we have ensured an even distribution of correct answers across options: 25.32\% in option A, 30.28\% in option B, 23.55\% in option C, and 20.83\% in option D. The dataset also incorporates application-based questions (approximately 18.84\%), such as given in Figure \ref{data_example}(left). Questions are stratified by difficulty: Level 1 (easy, 27.31\%), Level 2 (medium, 40.81\%), and Level 3 (hard, 31.86\%).

\noindent\textbf{Data Collection Process:}
The process, detailed in Figure \ref{pipeline}, represents the dataset creation process including that of intricate circuit visuals, as illustrated in Figure \ref{example}.

To construct the dataset, two students from a large public university in the U.S., both of whom had recently completed an undergraduate course in digital electronics covering the topics in the dataset, were enlisted as annotators. They independently formulated the questions based on the solved examples from resources 
\cite{lumetta2017ece120, feher},  
extracting or generating the corresponding images, and preparing the corresponding answers. About 80 visuals were extracted from textbooks, corresponding to 400 questions.
In all,  340 VQA instances were derived from course notes \cite{lumetta2017ece120}
and an additional 60 instances from textbook \cite{feher}, together covering a broad spectrum of undergraduate-level digital electronics concepts.

To ensure a comprehensive and balanced assessment across the diverse dimensions of digital electronics, we established three primary problem dimensions:

\textbf{Conceptual Dimension:} Key concepts fundamental to solving digital electronics problems were identified, including Karnaugh Map (K-map), Truth Table, Product of Sums (POS), Sum of Products (SOP), literal expressions, De Morgan’s theorem, area calculation, and gate delay. Additionally, to evaluate LLMs, we incorporated gate replacement, which involves substituting circuit elements with specific gates, and gate recognition, focusing on identifying gate types. These additions were informed by preliminary observations of the types of questions where models like ChatGPT showed deficiencies. This dimension holds multi-valued labels.

\textbf{Visual Context Dimension:} The dataset encompasses a variety of visual components such as finite-state machines (FSM), combinational gates, encoders/decoders, multiplexers/demultiplexers, flip-flops/latches, truth tables, transistors, clock diagrams, and K-maps. These elements represent the visual complexity inherent in digital electronics. This dimension holds multi-valued labels.

\textbf{Solving Strategy Dimension:} We categorized questions into three types: factual questions that require direct answers with no computation, computational problems that involve explicit and straightforward computational steps, and deep analytical questions that necessitate extensive domain knowledge and multiple reasoning steps, particularly in circuit optimization, trade-off evaluations and Application based questions. This dimension holds single-valued labels.

The distribution of these dimensions across the entire dataset is depicted in Figures \ref{fig4}, \ref{fig5}, and \ref{fig6}. 

\noindent\textbf{Synthetic Image Collection}
To enhance the dataset with more complex and diverse visuals beyond what textbooks typically offer, the annotators manually created 250 questions and answers, corresponding to 50 figures. Figures for these questions were then programmatically drawn using the \textit{schemdraw} library (\ref{schemdraw}).


\noindent\textbf{Data Review}
Following data collection, a rigorous review process was conducted by cross-referencing the questions between the two annotators, excluding answers. Annotators were permitted to use various resources, excluding large language models (LLMs), to solve these questions. Discrepancies in 65 questions were identified, leading to a 50\% discard rate after discussions. Additionally, questions deemed inadequate for evaluating LLM performance were removed. After resolving conflicts, the review process produced approximately 626 categorized questions from an initial pool of 800. Initially, we aimed to generate five textual questions per image, but maintaining this quantity compromised the quality of the questions. Further, the dataset's integrity was further verified by an instructor of a digital electronics course from a large university, ensuring its suitability for rigorous evaluation. More details provided in Appendix A.5\ref{sec:A.5}.

Although the questions and answers were manually curated to ensure high data quality, future efforts could benefit from automated recognition and extraction techniques for circuit diagrams, as seen in recent studies \cite{patare2016hand, thoma2021public, bayer2023text}. For instance, \cite{bayer2023text} proposed a method to extract textual information from hand-drawn circuit diagrams, which could significantly scale up the data curation process.

\begin{figure*}
  \begin{minipage}[t]{.32\linewidth}
    \includegraphics[trim = 0 20 0 10, width=\linewidth]{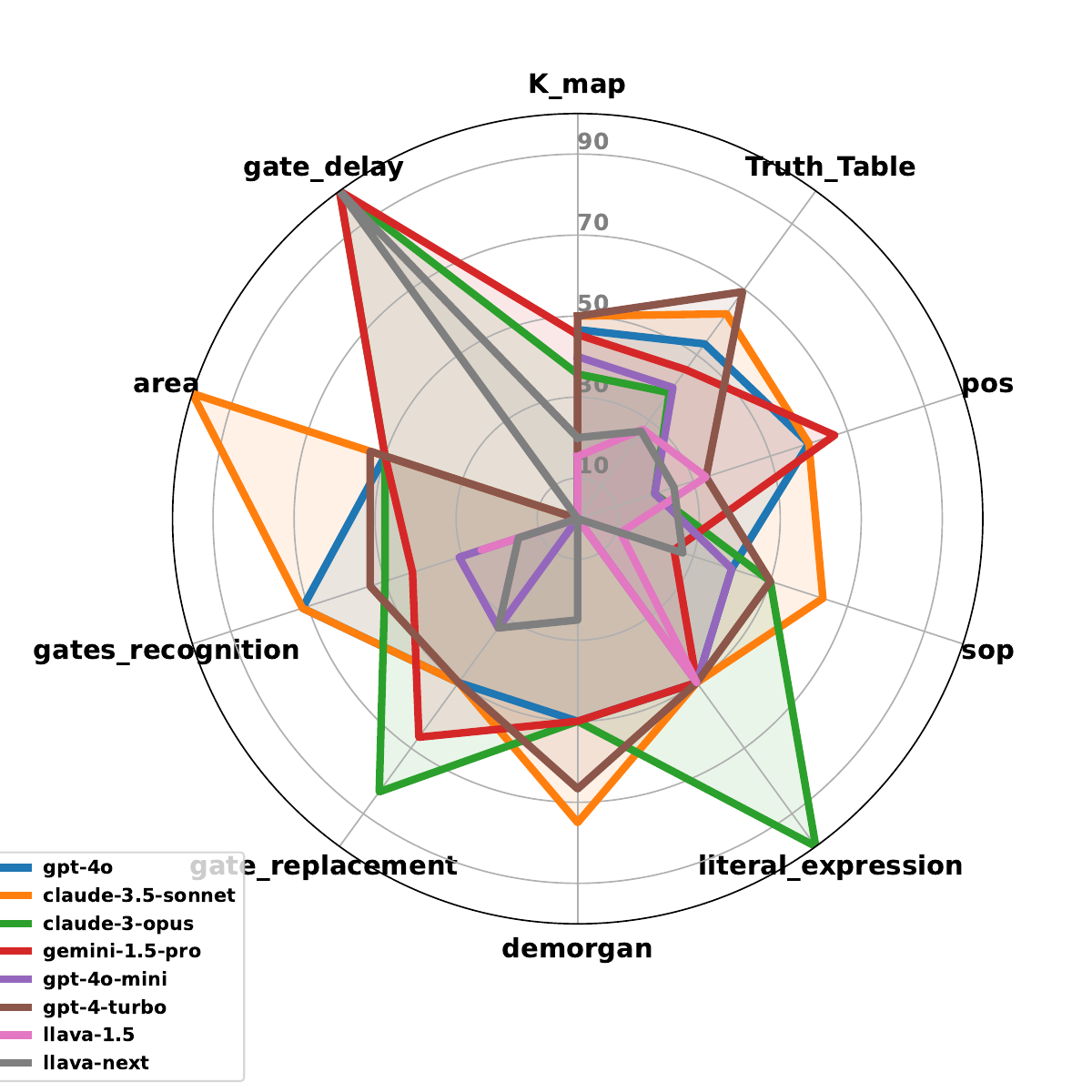}
    \subcaption{Basic concept}
    \label{fig1}
  \end{minipage}\hfil
  \begin{minipage}[t]{.32\linewidth}
    \includegraphics[trim = 0 20 0 10, width=\linewidth]{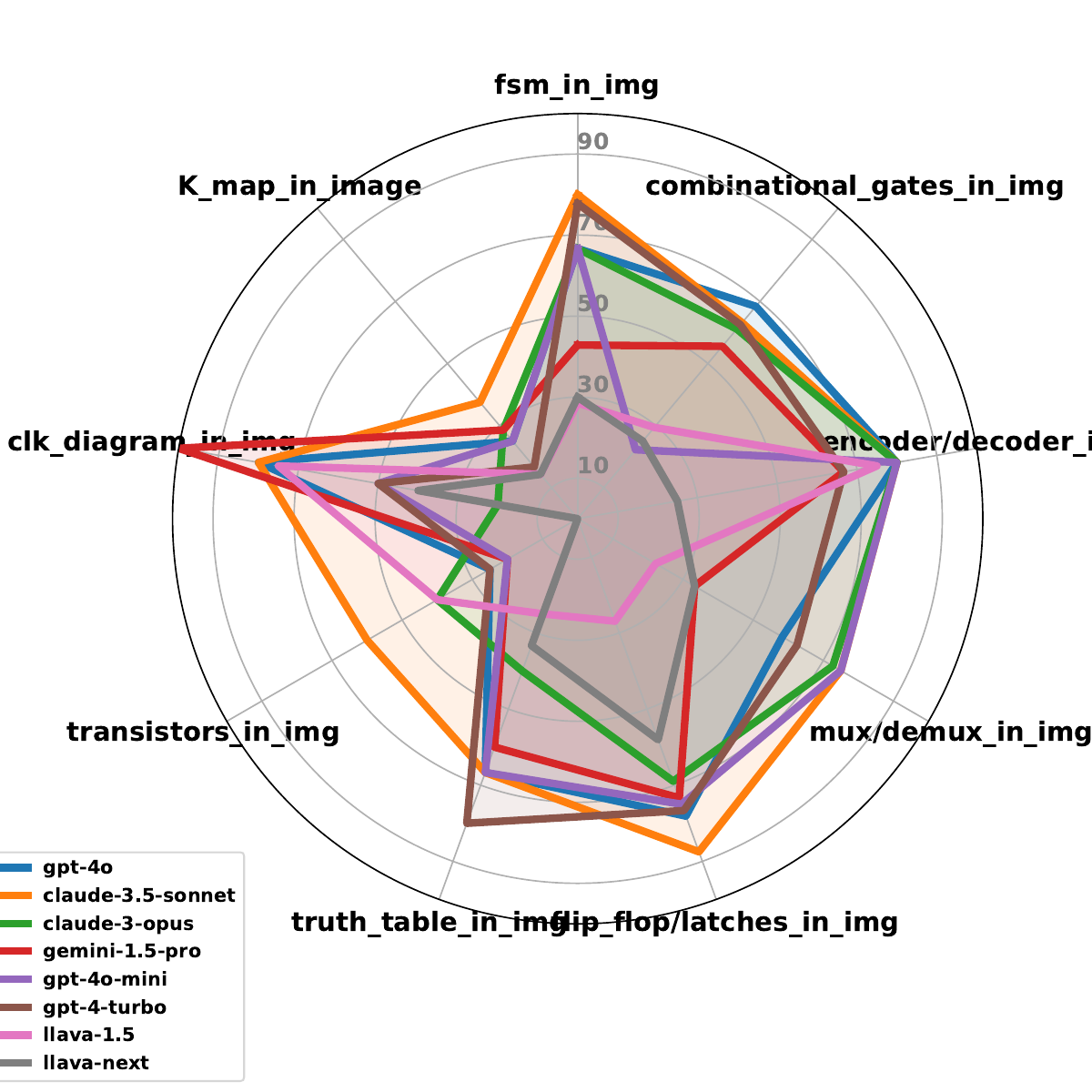} 
    \subcaption{Visual context}
    \label{fig2}
  \end{minipage}\hfil
  \begin{minipage}[t]{.32\linewidth}
    \includegraphics[trim = 0 20 0 10, width=\linewidth]{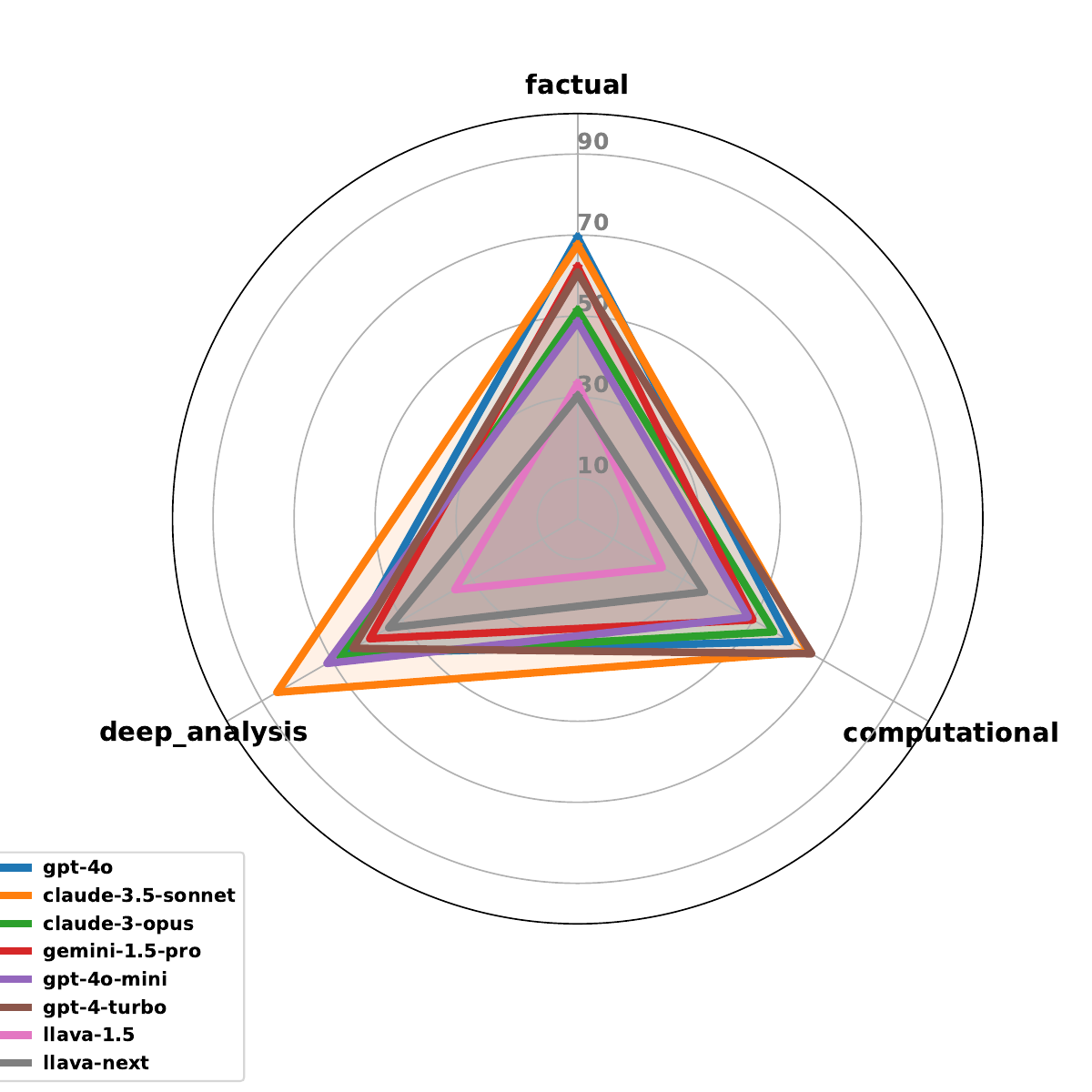}
\subcaption{Solving strategy}
\label{fig3}
  \end{minipage}%
  \caption{Performance of Large Language Models (LLMs) on a Visual Question Answering (VQA) task using CoT reasoning, across the categories on testmini 
  }
\end{figure*}

\section{Experimental Setup}
Our experiments are centered around answering the following research questions (RQ).\\
\textbf{RQ1.} How good are LLMs at answering the questions in our dataset using existing prompting methods? \textbf{RQ2.} Does LLMs' ability to answer these questions differ depending on whether visual or textual information was used? \textbf{RQ3.} What are the types of errors that LLMs make? \textbf{RQ4.} Can LLMs be leveraged to classify the error categories in their solutions? \textbf{RQ5.} What are the distribution and causes of errors made by LLMs?

\noindent\textbf{Experimental Data}
In addition to the full data, we closely evaluate the LLMs on a subset of our dataset, termed \textit{testmini}. This subset comprises 103 Visual Question Answering (\textbf{V}isual + \textbf{Q}uestion) instances. Of these questions, we had 57 with the \textit{Valid for Text Only} field enabled, meaning that those questions could be adequately answered only using the textual portion of the question combined with the \textit{Expression/Description} field, without relying on visuals. We call this grouping (\textbf{E}xpression + \textbf{Q}uestion). 
The distribution of all primary problem dimensions in this subset, and its comparison to that of the entire dataset, is shown in Figures \ref{fig4} to \ref{fig6}. 

\noindent\textbf{Metrics}
Since our questions have a single answer out of two or four choices, we use the accuracy of the final answer as the performance evaluation metric. \\
\noindent\textbf{Multi-modal Large Language Models}
We evaluate the proposed benchmark on several open-source multi-modal models, 
including Llava-1.5-7B \cite{liu2023llava} and Llava-Next-7B \cite{liu2023llava-next-interleave} known for their state-of-the-art visual-text processing capabilities. Subsequently, we assessed OpenAI's GPT-4o, GPT-4o-mini, and GPT-4-turbo~\cite{achiam2023gpt} which are strong at generalization across multi-modal tasks, along with emerging multi-modal models such as Gemini-1.5-pro~\cite{reid2024gemini}, and the Claude models~\cite{anthropic2024claude}. For the (E+Q) questions, we also assessed Meta's language models Llama-3-70B-Instruct and Llama-3.1-405B \cite{touvron2023llama3}. 

To obtain the model's response, each model is prompted with the expected response type formatted as a multiple choice question, concatenated with the problem description and either the visual component or the \textit{Expression/Description} of the question with a maximum token limit of 600, because most instances were within this range.
In addition, we investigate the zero-shot Chain-of-Thought (CoT) \cite{wei2022chain} prompting technique.
Given the advanced capabilities of GPT-4o, the exact answer was extracted by re-prompting GPT-4o with the responses generated by the various MLLMs.
If an LLM's response was incoherent or the expected response type did not match any available options, the response was recorded as ``None."

\begin{table}[h!]
\centering
\resizebox{\columnwidth}{!}{
\begin{tabular}{c|cc|cccc}
\multicolumn{1}{l|}{} & \multicolumn{2}{c|}{\textbf{\begin{tabular}[c]{@{}c@{}} Testmini\\ 103 samples \end{tabular}}}&\multicolumn{4}{c}{\textbf{\begin{tabular}[c]{@{}c@{}} \textit{Valid for text only} = 1\\ 57 samples \end{tabular}}} \\ \hline
\textbf{LLMs} & \multicolumn{1}{c}{\textbf{\begin{tabular}[c]{@{}c@{}}- CoT\\ V+Q\end{tabular}}} & \multicolumn{1}{c|}{\textbf{\begin{tabular}[c]{@{}c@{}} + CoT\\ V+Q\end{tabular}}} & \multicolumn{1}{c}{\textbf{\begin{tabular}[c]{@{}c@{}}- CoT\\ E+Q\end{tabular}}} & \multicolumn{1}{c}{\textbf{\begin{tabular}[c]{@{}c@{}}+ CoT\\ E+Q\end{tabular}}} & \multicolumn{1}{c}{\textbf{\begin{tabular}[c]{@{}c@{}}- CoT\\ V+Q\end{tabular}}} & \multicolumn{1}{c}{\textbf{\begin{tabular}[c]{@{}c@{}}+ CoT\\ V+Q\end{tabular}}} \\ \hline
\textbf{GPT-4o} & 67.38  & 66.33 & \textbf{75.0}  & 74.0 &\textbf{ 70.90} & \textbf{64.91}\\
\textbf{GPT-4o-mini}     & 50.0    & 55.0    & 71.69   & 72.22 & 47.27  & 48.21 \\
\textbf{GPT-4-turbo}     &  62.5   &  63.63  &  \textbf{ 75.0} &  74.0 &68.42 & 61.53\\
\textbf{Claude-3-opus}     & 60.20  & 57.73 & 58.69  & 55.55 &60.00 &57.44 \\
\textbf{Claude-3.5-sonnet }     & \textbf{69.30}   & \textbf{71.56} & 55.31 & 57.44  &60.00& 57.44\\
\textbf{Gemini-1.5-pro}     & 56.12  & 57.44   & 73.33  & 72.10     & 55.55 &56.60  \\ \hline
\textbf{Llava-Next} & 40.81 & 38.94    & 48.21 & 46.29  &42.85& 37.25 \\
\textbf{Llava-1.5}  & 35.57                                                                               & 29.62                                                                              & 39.21                                                                             & 40.0   &29.82 &  29.16                                                                    \\ 
\textbf{Llama-3-70B-Instruct}  & x  & x & 59.64  & \textbf{75.0}  &x&x                                                                \\
\textbf{Llama-3.1-405B}  & x  & x &  71.69 &   63.63  &x&x  \\
\textbf{Average Performance} & 54.28 & 54.36 & 62.46 & 63.79 & 53.39 & 50.65 \\ 
\end{tabular}}
\caption{Comparison of various LLMs' performance with and without Chain-of-Thought (CoT) prompting for visual and text-only question answering using expressions in digital images on testmini}
\label{table1}
\end{table}

\subsection{Results and Discussions}
We now present the major findings of our experiments.
\subsubsection{RQ1. How good are LLMs at answering the questions in our dataset using existing prompting methods?} 
Table~\ref{table1} compares various MLLMs, both with and without Chain-of-Thought prompting, for visual (V+Q) and text-only (E+Q) question answering. Our findings indicate that open-source models generally trail behind closed-source ones in VQA performance. The average accuracy of LLMs on entire testmini dataset is approximately 54\%, highlighting the challenges these questions pose. Notably, Claude-3.5-sonnet outperformed all the other models in VQA tasks on the testmini data which includes additional challenging questions that cannot be represented by simple expressions. Figures \ref{fig1}, \ref{fig2}, and \ref{fig3} show that Claude-3.5-sonnet consistently led across the three primary problem dimensions of the V+Q data. A closer look at per-category performance reveals specific challenges: LLMs often struggled with transistor image-based questions under the visual context dimension (Figure~\ref{fig2}), and area calculations  were frequently incorrect across most models, suggesting model capabilities in these subject areas need further improvement. 

Interestingly, models performed relatively well in the deep analysis category (Figure \ref{fig3}), despite the questions being more challenging.
This could indicate some level of memorization of conceptual statements by the models, warranting further investigation.
Additionally, CoT prompting did not consistently improve performance in our task, unlike its effectiveness in other problem-solving scenarios \cite{wei2022chain, arora2023have}. 
\noindent \textbf{Evaluation on complete dataset:}
Having analyzed LLMs' performance on testmini, we investigate the performance of the two leading models on the full dataset and observe similar trends as seen on testmini in Table \ref{tableleading}.
GPT4-o achieved an accuracy of 60.39\%, while Claude-3.5-sonnet reached 62.27\% on the (V+Q) dataset. However, Claude's performance significantly declined to 45.21\% accuracy on the (E+Q) dataset, whereas GPT4-o's performance surged, achieving 72.36\% accuracy. Further analysis is provided in Appendix A.4\ref{sec:A.4}.

\subsubsection{RQ2. Does LLMs' ability to answer these questions differ depending on whether visual or textual information was used?}
In Table~\ref{table1} 
Claude-3.5-sonnet model showed the best performance for V+Q on the full testmini dataset, while GPT-4o and GPT-4 were best among all LLMs for the \textit{Valid for text-only equal to 1} subset in both (V+Q) and (E+Q) evaluations. This indicates that closed models outperform open-sourced models on our dataset. \textit{Llama3}, on the other hand, exhibits comparable efficacy in the E+Q setting showing the promise of open-source \textit{language} models, but there is a lack of similarly powerful open-source MLLMs for handling the visual components. Additionally, most of these models' effectiveness declines sharply in the V+Q setting compared to that of the E+Q; only Claude models show the opposite trend. This suggests that most models struggle with complex visual understanding on our dataset. 

\subsubsection{RQ3. What are the types of errors that LLMs make?}
Given the limited abilities of GPT-4o and other models on our benchmark, error detection and correction are critical steps in addressing the remaining inaccuracies in solutions generated by MLLMs. To this end, we conduct an extensive manual error analysis by manually grouping the errors into broad types by the annotators who created the questions. Since a model can generate the final correct answer despite making some errors in the response, we analyze the complete responses for all samples, not just ones with incorrect final answers. Therefore, we leverage our manual error analysis to closely analyze the performance of the two leading models, namely Claude and GPT-4o, and investigate their textual vs. visual understanding capabilities on our dataset below. 
\begin{figure}[t]
\centering
\includegraphics[trim=10 40 10 25, width=0.9\linewidth]{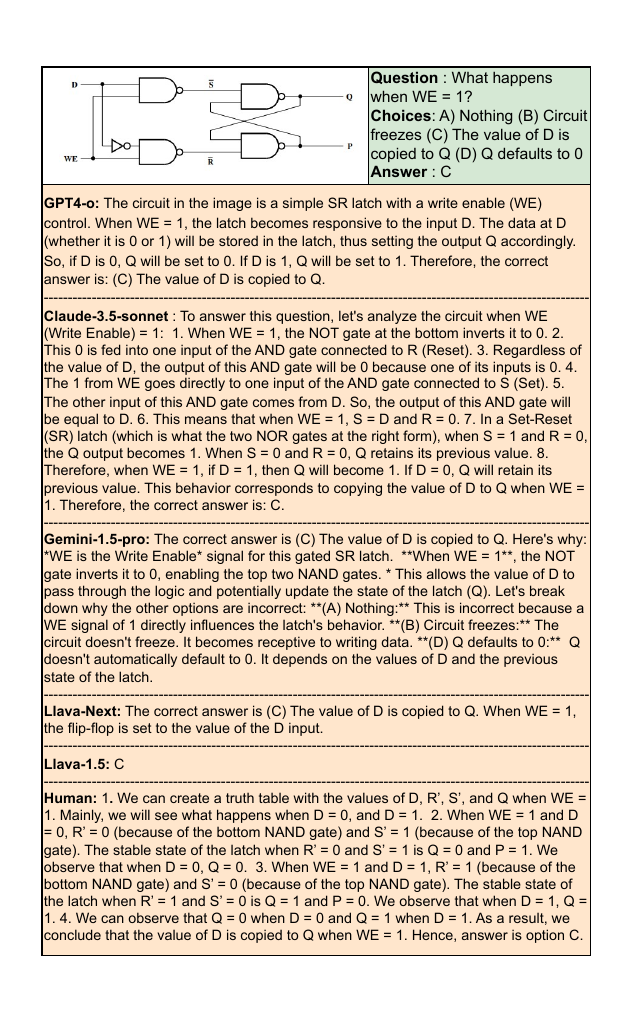} 
\caption{MLLM's responses for the sample question}
\label{llm_response}
\end{figure}

\begin{figure*}
 \begin{minipage}[t]{.32\linewidth}
    \includegraphics[trim = 0 30 0 40 , width=\linewidth]{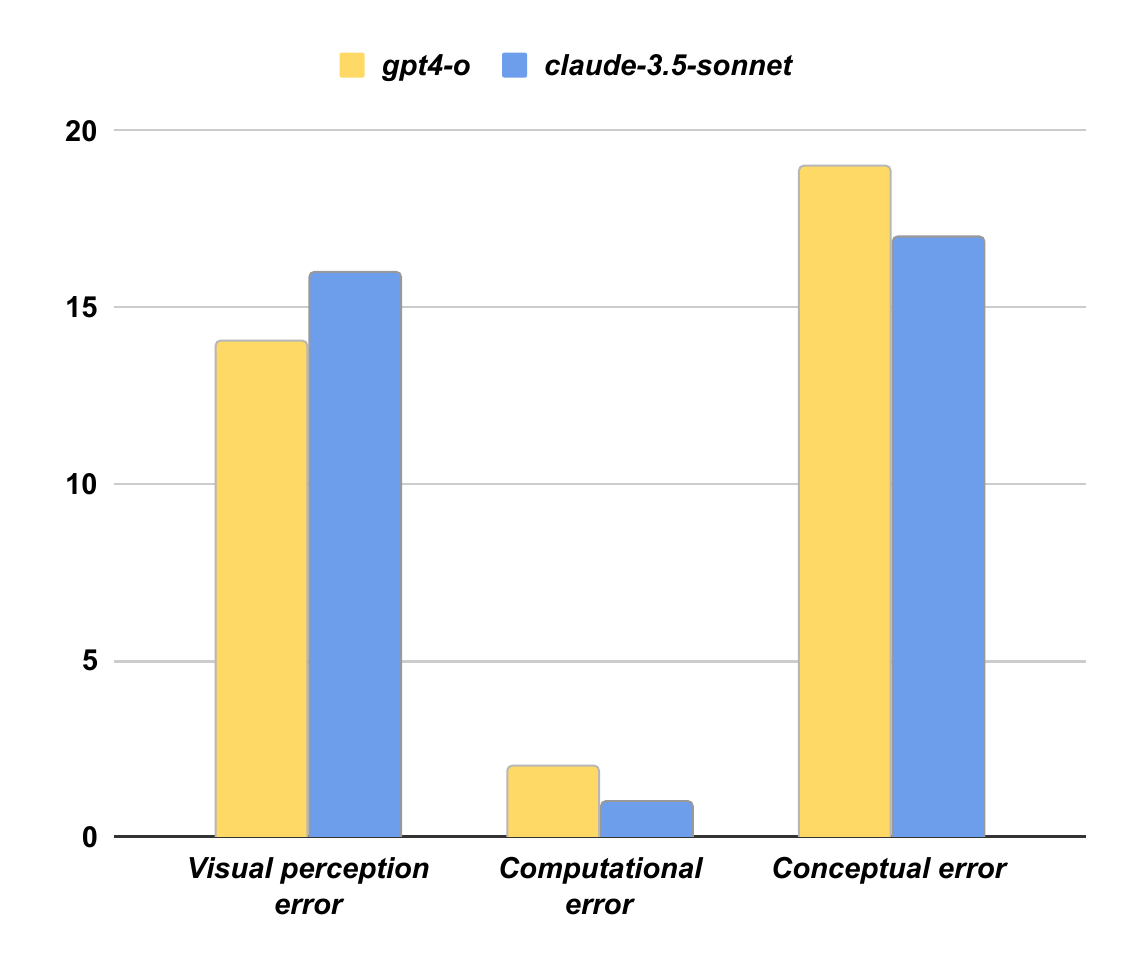} 
    \subcaption{}
    \label{gptvsclaude}
  \end{minipage}\hfil
  \begin{minipage}[t]{.32\linewidth}
    \includegraphics[trim = 0 30 0 40 ,width=\linewidth]{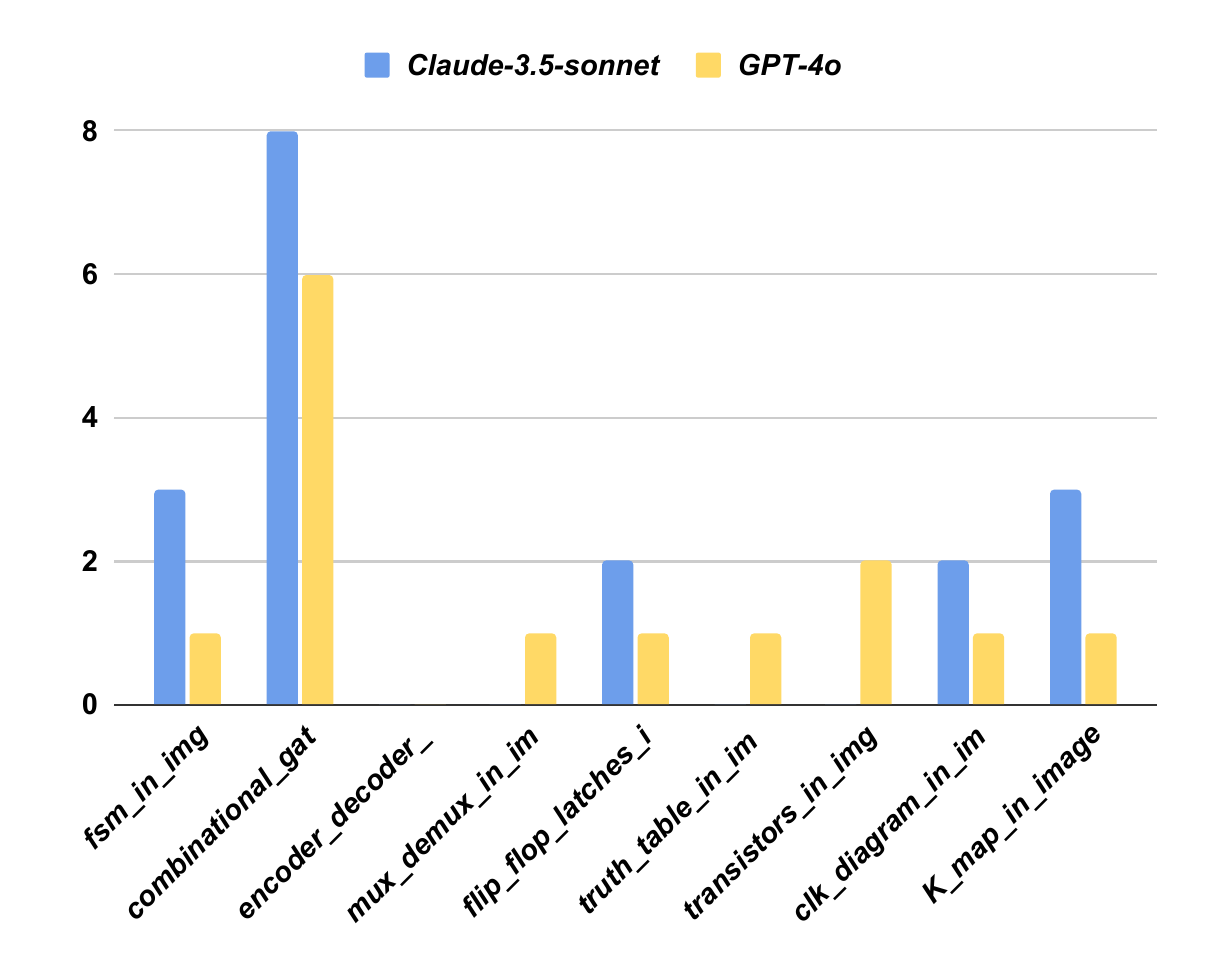}
    \subcaption{}
    \label{visual_perception}
  \end{minipage}\hfil
  \begin{minipage}[t]{.32\linewidth}
    \includegraphics[trim = 0 30 0 40 ,width=\linewidth]{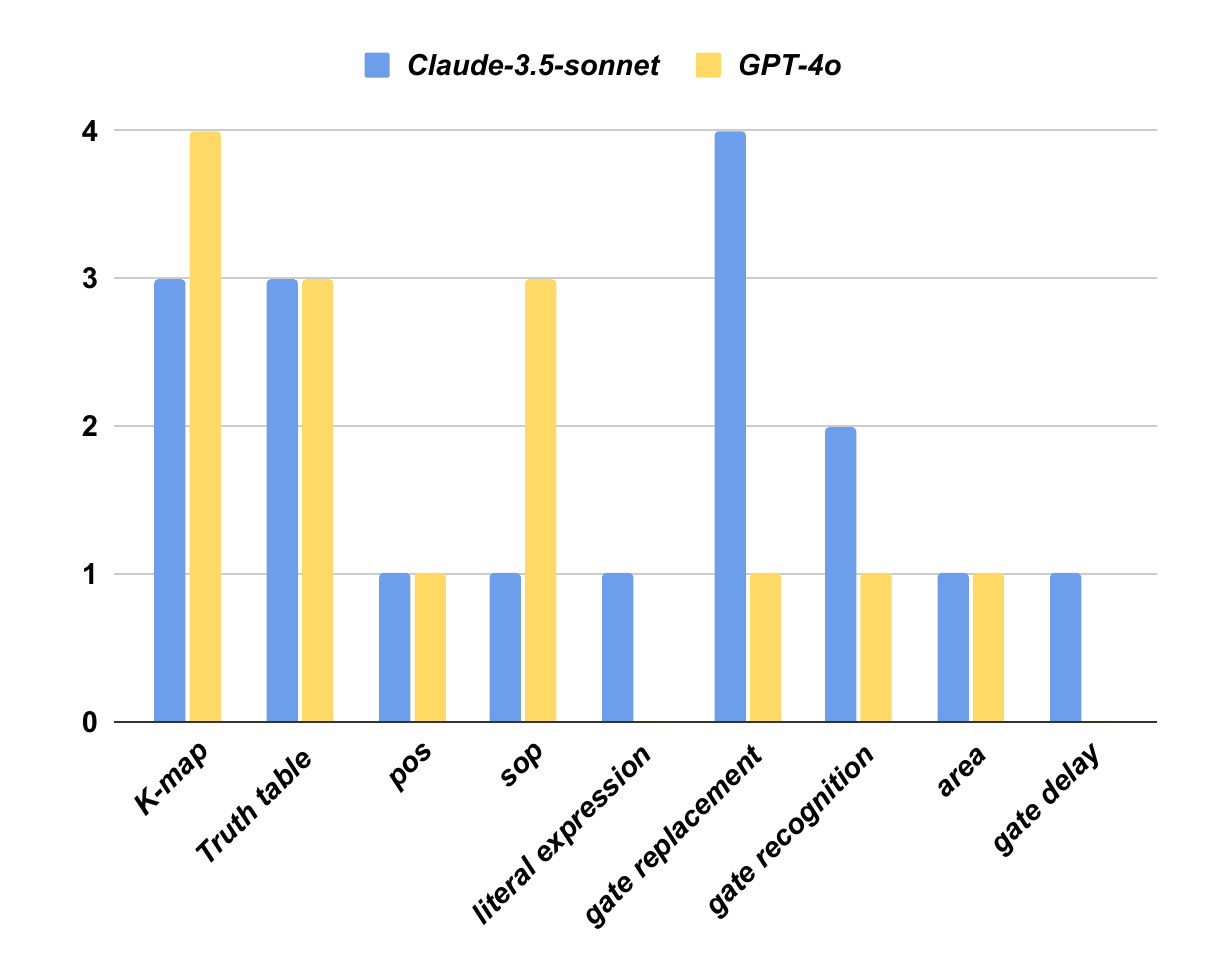}
\subcaption{}
\label{Conceptual}
  \end{minipage}%
  \caption{(a): The error counts made by GPT-4o and Claude for (V+Q) data, utilizing 98 samples, (b): Visual perception error distribution based on visual entity, which was traced in (a). (c): Conceptual error distribution based on concepts used while solving questions, which was traced in (a)}
\end{figure*}
Our analysis identified four predominant types of errors described below.\\
\textbf{Problem comprehension error:} Failure to understand the textual problem correctly. \\
\textbf{Visual perception error}: Error arises when there is a misinterpretation of entities within the image, especially during perception. It typically can occur when all steps for the associated textual question are correct but the final answer is wrong. \\
\textbf{Computational Error}: This error generally occurs when there is a mistake in calculations or algorithm execution, resulting in incorrect outputs. \\
\textbf{Conceptual Error}: This error arises from misunderstanding or misapplication of a concept after perceiving the correct information through images and other details. \\
Examples of these categories provided in Appendix A.3 \ref{sec:A.3}.
\subsubsection{RQ4. Can LLMs be leveraged to classify the error categories in their solutions?} 
We explore the possibility of leveraging GPT-4o to automatically detect those errors in the steps generated by various models. For this, inspired by Chain of Thought (CoT) evaluation strategy\cite{zhang2024mathverse}, we design a CoT prompt that asks GPT-4o to identify the category of error (if any) in the model-generated response.  
The prompts utilized for this strategy are detailed in Appendix A.2 \ref{sec:A.2}.
As illustrated in Figure \ref{manualvsprompt}, this strategy revealed a significant discrepancy between the errors detected by prompts and those identified through manual annotation. Specifically, the number of errors detected by the prompt-based method was significantly lower than what could be identified manually, suggesting that this strategy may overlook a considerable number of errors. These results underscore the need for further refinement in prompt-based error detection techniques to more closely align with human judgment and enhance the robustness of MLLM outputs.

\begin{figure}[h!]
\centering
\includegraphics[trim = 10 55 10 30, width=0.9\columnwidth]{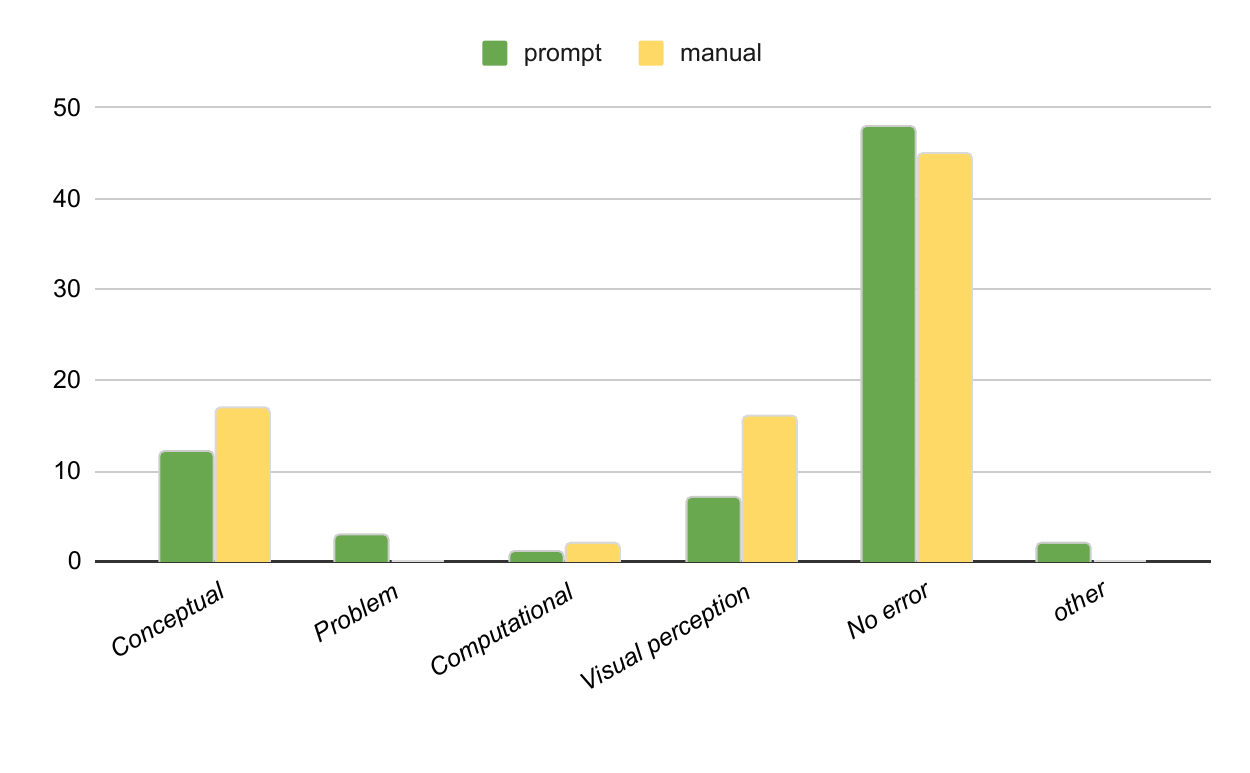} 
\caption{Comparison of errors detected by GPT-4o vs. manual categorisation for 98 samples with \textit{non-None} answers}
\label{manualvsprompt}
\end{figure}


\subsubsection{RQ5: What are the distribution and causes of errors made by LLMs?}


As shown in Table \ref{table1}, the performance of leading models, GPT-4o and Claude is still lacking.
To investigate this issue, we conduct a detailed analysis of manually categorized errors across the primary problem dimensions defined for about 98 (V+Q) question-answers with non-None entries. 
In Figure \ref{gptvsclaude}, both models exhibit significantly higher error rates in visual perception and conceptual understanding compared to computational and problem comprehension errors. This further underscores that while these LLMs excel at processing textual information and performing binary math, they struggle with visual and conceptual tasks.

Further,
Figures \ref{visual_perception} and \ref{Conceptual},
show that GPT-4o's poor performance is due to a weak understanding of gates from visuals has been validated, and this issue is also evident in Claude. Conceptual errors primarily arise in gate replacement questions, which demand deeper analysis.  
Additionally, both models show deficiencies in handling truth tables and K-map-based concepts.  To mitigate one major category errors namely gate recognition, we did a preliminary exploration of prompt-tuning but did not notice improvements as shown in Appendix A.1\ref{sec:A.1}. Thus, further research is needed to improve MLLM capabilities in solving questions about these concepts.


\subsubsection{Textual vs. visual understanding of Claude and GPT-4o}
\begin{table}[h!]
\centering
\resizebox{\columnwidth}{!}{
\begin{tabular}{c|cccc}
\textbf{Error type}      & \multicolumn{1}{c}{\textbf{\begin{tabular}[c]{@{}c@{}} GPT-4o \\ (E+Q)\end{tabular}}}  & \multicolumn{1}{c}{\textbf{\begin{tabular}[c]{@{}c@{}} GPT-4o \\ (V+Q)\end{tabular}}} & \multicolumn{1}{c}{\textbf{\begin{tabular}[c]{@{}c@{}} Claude-3.5-sonnet \\ (E+Q)\end{tabular}}} & \multicolumn{1}{c}{\textbf{\begin{tabular}[c]{@{}c@{}} Claude-3.5-sonnet\\ (V+Q)\end{tabular}}}\\ \hline
Conceptual  & 4 & 11& 11 & 12\\
Computational     & 2  &  2 &   1 &  1 \\
Problem comprehension     & 2  & 0  & 5 &  0 \\
Visual perception & 0 &  8 &0 &  12\\
\end{tabular}}
\caption{The error counts made by GPT-4o and Claude in about 47 text-only (non-None responses) (E+Q) and (V+Q) question answering, utilizing expressions/descriptions and digital images respectively 
}
\label{table5}
\end{table}



To compare the visual and textual understanding abilities of the respective LLMs on our data, we categorized the errors made by both models in both (E+Q) and (V+Q) as shown in Table \ref{table5}. For (E+Q), Claude demonstrates deficiencies, particularly in conceptual errors and problem comprehension errors. However, in (V+Q), both the LLMs show difficulties perceiving from visuals. Although the performance of Claude with CoT prompting in Table \ref{table1} is comparable in both (E+Q) and (V+Q) settings, its error counts in those settings suggests that it has provided the correct answers despite making visual perception errors. Overall, this underscores the poor capabilities of MLLMs in understanding electronics diagrams in our dataset.

\section{Conclusion and Future Work}

Our analysis reveals that LLMs like GPT-4o excel in language tasks but struggle with Visual Question Answering (VQA) in digital electronics, particularly with basic digital gates. Errors from GPT and Claude models highlight the specific deficiencies, raising the question of how to enhance their capabilities. Enhancing LLM capabilities could involve integrating online solvers and electronic design tools, though such resources are currently limited. To advance the field, we suggest focusing on multimodal data processing in LLMs, improving foundational engineering understanding, and using our new ElectroVizQA benchmark to guide future research and address these limitations.
\bibliography{aaai25}

\clearpage
\appendix
\section{Appendix A.1}\label{sec:A.1}
\subsection{Prompting techniques}
\subsubsection{Prompt used for evaluation on {E+Q}} 
The following prompt was employed to evaluate LLMs on the (E+Q) set. Given that the textual component of the questions is primarily designed for visuals in (V+Q) type questions, we provided the contents of images in the prompts. 
\begin{tcolorbox}[colback=gray!5!white,colframe=gray!75!black, colbacktitle=gray!75!black,title= Evaluation on (E+Q) prompting ]
Given a figure representing \{expression\}, solve the following question 
\{question\_format\}
\end{tcolorbox}
We followed the following format to feed questions to various LLMs.\\
\textbf{Question\_format} = Please answer the question and provide the correct option letter, e.g., A, B, C, D at the end. \textit{Question:} What happens when S' has a value of 1? \textit{Choices:} (A) No effect on dual inverter loop (B) Loop becomes automatically unstable (C) Loop becomes automatically stable
\subsubsection{Answer extraction}
For the Visual Question Answering (VQA) task, we employed a multi-stage prompting strategy. Initially, we extracted responses from each LLM by simply providing the question along with a hint about the expected answer format, as the questions were single-correct multiple-choice questions. After obtaining the initial responses, we then used GPT-4o to refine these by extracting the option choices through advanced prompting techniques. This approach allowed us to systematically parse and evaluate the LLMs' answers, ensuring that the extracted choices aligned with the format and structure required for accurate assessment. 

\begin{tcolorbox}[colback=gray!5!white,colframe=gray!75!black, colbacktitle=gray!75!black,title= Answer extraction prompt]
I am providing you a response from a model to a digital circuit problem, under 'Model Response'. You should extract the answer from the response as 'Extracted Answer'. Directly output the extracted answer with no explanation, just one letter output.

1.\\
\textbf{Model response:} "The provided circuit is a NAND gate. The function of a NAND gate is the complement of the AND function. The logical expression for a NAND gate with inputs A and B is: [C = A NAND B] To convert this into Sum of Products (SOP) form, we use De Morgan's theorem: [A NAND B = A' + B'] Therefore, the equivalent SOP expression for this circuit is: [A' + B']. The correct option is (C)."\\
\textbf{Extracted Answer: }C\\
2.\\
\textbf{Model response:} "The provided circuit is a NAND gate. The function of a NAND gate is the complement of the AND function."\\
\textbf{Extracted Answer:} None\\
\textbf{Model response:} {response}\\
\textbf{Extracted Answer:}
\end{tcolorbox}
\subsubsection{Textual prompting to improve gate recognition accuracy}
We attempted several methods to enhance the basic gate recognition performance of LLMs, particularly GPT-4o. Initially, we included detailed textual descriptions of gate appearance and structure in the prompts to guide the model's understanding. Despite using the precise prompt provided below, the LLMs consistently returned incorrect answers, indicating that the added descriptions were insufficient to improve recognition accuracy.

\begin{tcolorbox}[colback=gray!5!white,colframe=gray!75!black, colbacktitle=gray!75!black,title= Prompt provided to improve gate detection accuracy]
AND Gate: A flat-ended rectangle with a curved, semi-circular right side, distinguishing it by its flat input side and rounded output side. \\
NAND Gate: An AND gate symbol with a small circle (inversion bubble) at the output, distinguishing it by the inversion bubble indicating negation. \\
OR Gate: A curved shape that converges to a point on the right side, distinguishing it by its pointed output side and curved input side. \\
NOR Gate: An OR gate symbol with a small circle (inversion bubble) at the output, distinguishing it by the inversion bubble indicating negation. \\
NOT Gate: A triangle pointing to the right with a small circle (inversion bubble) at the output, distinguishing it by its single input and the inversion bubble. \\
XOR Gate: An OR gate symbol with an additional curved line inside and parallel to the input side, distinguishing it by the extra inner curved line representing exclusivity. \\
XNOR Gate: An XOR gate symbol with a small circle (inversion bubble) at the output, distinguishing it by the combination of the XOR shape and the inversion bubble indicating negation. \\
Please answer the question and provide the correct option letter, e.g., A, B, C, D at the end. \\
Question: K is the direct output of which type of gate? \\
Choices: (A) XOR gate (B) XNOR gate (C) NAND gate (D) NOR gate
\end{tcolorbox}
\subsubsection{Visual prompting to improve gate recognition accuracy}
In this approach, we provided visual prompts that included gate names alongside their corresponding visuals, aiming to enhance the LLMs' gate recognition capabilities. Despite this, the method proved ineffective. Even after making various edits to the visual prompts such as highlighting the distinguishing features of each gate—the LLMs continued to struggle with accurate recognition. This suggests that simply presenting visual cues, even with enhanced features, is insufficient for improving performance in gate identification tasks.
\section{Appendix A.2}\label{sec:A.2}
\subsection{Extended (CoT) Strategy for Error Type Recognition}
We experimented with prompting methods by providing GPT-4o with an extended version of the CoT strategy proposed by \cite{zhang2024mathverse}. This approach aimed to improve error categorization, aligning more closely with manual methods, as shown in Figure \ref{manualvsprompt}. However, our results indicate that this method is unreliable, with a significant number of responses incorrectly categorized. Additionally, we observed inconsistencies in the calculation of average and final scores, which were sometimes inaccurate or buggy.
\begin{tcolorbox}[enhanced,attach boxed title to top center={yshift=-3mm,yshifttext=-1mm},
  colback=gray!5!white,colframe=gray!75!black,colbacktitle=gray!80!black,
  title=Extended (CoT) Strategy for Error Type Recognition, fonttitle=\bfseries,
  boxed title style={size=small,colframe=gray!50!black} ]
  I will provide you with a visual digital electronics problem, including the question, diagram, and ground-truth answer, and then give you a model output containing multiple key solution steps.
Please think step by step and output the Average score, along with the Final answer score and the error type in the end, as described below\\
\textbf{Average score}: Evaluate, based on the given question, answer, and diagram whether each solution step is correct in understanding the question's objective, visual perception, and logical computation, with an incorrect score of 0 and a correct score of 1. Then, calculate the average score of multiple steps.\\
\textbf{Final answer score}: Match the Model's extracted option with the Ground-truth answer, scoring 1 if it matches and 0 if it doesn’t. If the model output only includes a single step or answer, the Average score and Final answer score are the same. \\
\textbf{Problem miscomprehension error}: Failure to understand the textual problem correctly. \\
\textbf{Visual perception error}: Error arises when there is a misinterpretation of entities within the image, especially during perception. It typically can occur when all steps for the associated textual question are correct, and fully match with one of the choices provided but the final answer is wrong.\\
\textbf{Conceptual error}: Error arises on misunderstanding or misapplication of a concept after perceiving the information through image and other details.\\
\textbf{Computational error}: Mistake in calculations or algorithm execution, resulting in incorrect outputs. \\
\textbf{Error type}: Based on the above definition error type can be  "Conceptual error ", "Computational error ", "Visual perception error", "Problem Miscomprehension error" or "No error".
At the end provide the final answer for Average score, Final answer score, and Error type only without any explanation\\
\textbf{Question}: {question}\\
\textbf{Ground-truth answer}: {correct\_option}\\
\textbf{Model's extracted option}: {model\_extracted\_option}\\
\textbf{Model's extracted output}: {model\_extracted\_steps}\\
\textbf{Average score, Final answer score, Error type}: 
\end{tcolorbox}

\section{Appendix A.3} \label{sec:A.3}
\subsection{Examples of different Error types}
\begin{tcolorbox}[colback=gray!5!white,colframe=gray!75!black, colbacktitle=gray!75!black,title= Problem miscomprehension error]
    \begin{minipage}{0.3\textwidth} 
        \centering
        \includegraphics[width=\textwidth]{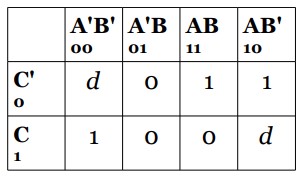} 
    \end{minipage}%
    \hfill
    \begin{minipage}{0.65\textwidth} 
        \textbf{Question:} How many terms does the minimal SOP expression have?\textbf{ Choices:} (A) 0 (B) 1 (C) 2 (D) 3 \\
        \textbf{Correct answer:} C
    \end{minipage}
    \tcblower
   Let's think step by step:\\
   1. \textit{\textbf{The minimal SOP expression has only one term. (Wrong interpretation)}} \\2. The expression is a binary number, which means it has only two possible values: 0 or 1.\\3. The expression is a one-bit binary number, which means it has only one bit (0 or 1) and no other terms.\\ Therefore, the correct answer is (B) 1
\end{tcolorbox}
\begin{tcolorbox}[colback=gray!5!white,colframe=gray!75!black, colbacktitle=gray!75!black,title= Computational error]
    \begin{minipage}{0.2\textwidth} 
        \centering
        \includegraphics[width=\textwidth]{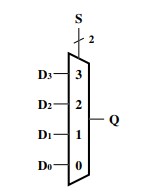} 
    \end{minipage}%
    \hfill
    \begin{minipage}{0.7\textwidth} 
        \textbf{Question:} What is the equivalent expression for this mux?  \textbf{Choices:} (A)  S1'S0'D1 + S1'S0D2 + S1S0'D3 + S1S0D4  (B)  S1S0D1 + S1'S0D2 + S1S0'D3 + S1S0D4  (C)  S1'S0'D1 + S1'S0D2 + S1S0D3 + S1S0D4  (D)  S1'S0'D1 + S1'S0'D2 + S1'S0'D3 + S1S0D4 \\
        \textbf{Correct answer:} A
    \end{minipage}
    \tcblower
    To determine the equivalent expression for the given multiplexer (MUX), we need to carefully analyze how the select inputs (S1) and (S0) control the output (Q) based on the data inputs (D3, D2, D1, D0). Given the 4-to-1 multiplexer: \\
    - When (S = 00), output(Q = D0) \\
    - When (S = 01), output (Q = D1) \\
    - When (S = 10), output (Q = D2) \\
    - When (S = 11), output (Q = D3) \\ 
    In terms of binary selection inputs (S1S0): \\
    (S1'S0') selects (D0), (S1'S0) selects (D1), (S1S0') selects(D2), (S1S0) selects (D3)\\
    To construct the expression for (Q), we'll use the complemented and uncomplemented forms of the select inputs along with the data inputs: Q = S1'S0'D0 + S1'S0D1 + S1S0'D2 + S1S0D3 
    Simplifying it for the choice format given in the question:\\
    \textbf{\textit{Q = S1'S0'D1 + S1'S0D2 + S1S0'D3 + S1S0D4
    Thus, the correct choice matches the expression:
    (C) ( S1'S0'D1 + S1'S0D2 + S1S0D3 + S1S0D4 ) (Wrong comparison with choices)}}
    So, the correct option is C.
\end{tcolorbox}
\begin{tcolorbox}[colback=gray!5!white,colframe=gray!75!black, colbacktitle=gray!75!black,title= Visual perception error]
    \begin{minipage}{\textwidth} 
        \centering
        \includegraphics[width=\textwidth]{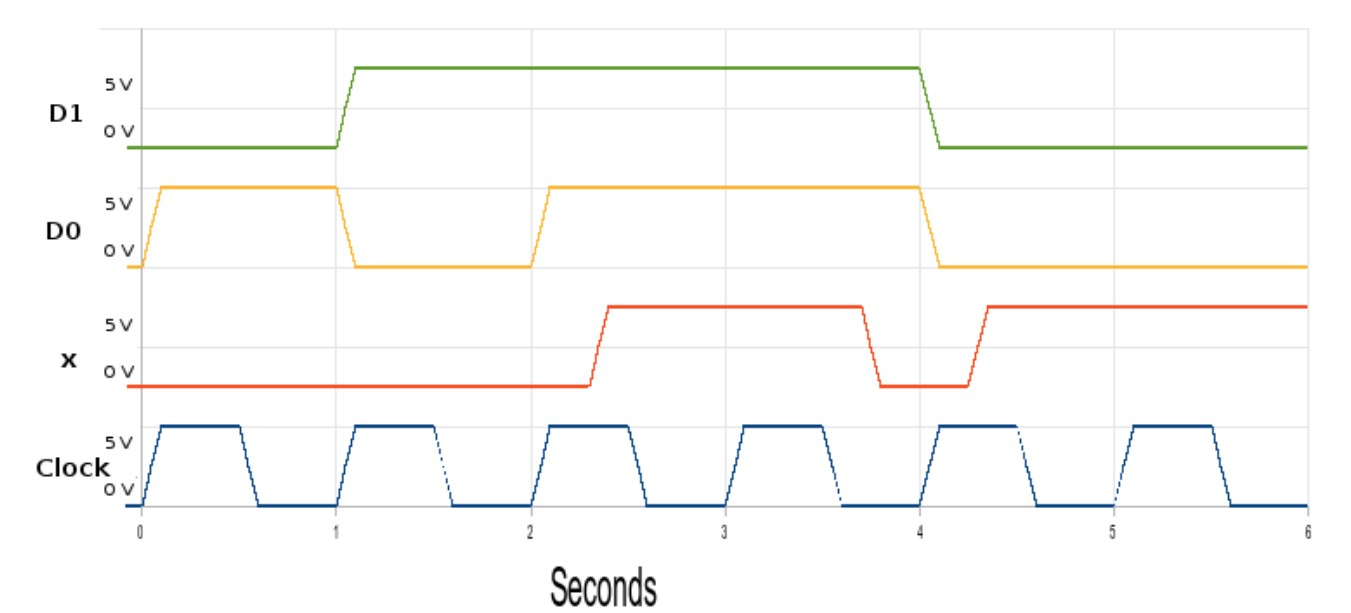} 
    \end{minipage}%
    \hfill
    \begin{minipage}{\textwidth} 
        \textbf{Question:} True or False? x only changes during the rising edge of the clock signal.\\
        \textbf{ Choices:} (A) True (B) False \\ \textbf{Correct answer:} B
    \end{minipage}
    \tcblower
   Let's analyze the waveform diagram to determine when x changes relative to the clock signal. 1. Examine the signal x and the clock signal closely. 2. Notice the points at which x changes between 0V and 5V. 3. Compare these change points to the clock signal's rising and falling edges.\\ From the diagram: - \textbf{\textit{x transitions from 0V to 5V and from 5V to 0V are synchronized with the rising edge of the clock signal. (Wrong perception)}}\\ Therefore, the statement x only changes during the rising edge of the clock signal is correct. \\ Answer: (A) True
\end{tcolorbox}
\begin{figure*}
  \begin{minipage}[t]{.32\linewidth}
    \includegraphics[width=\linewidth]{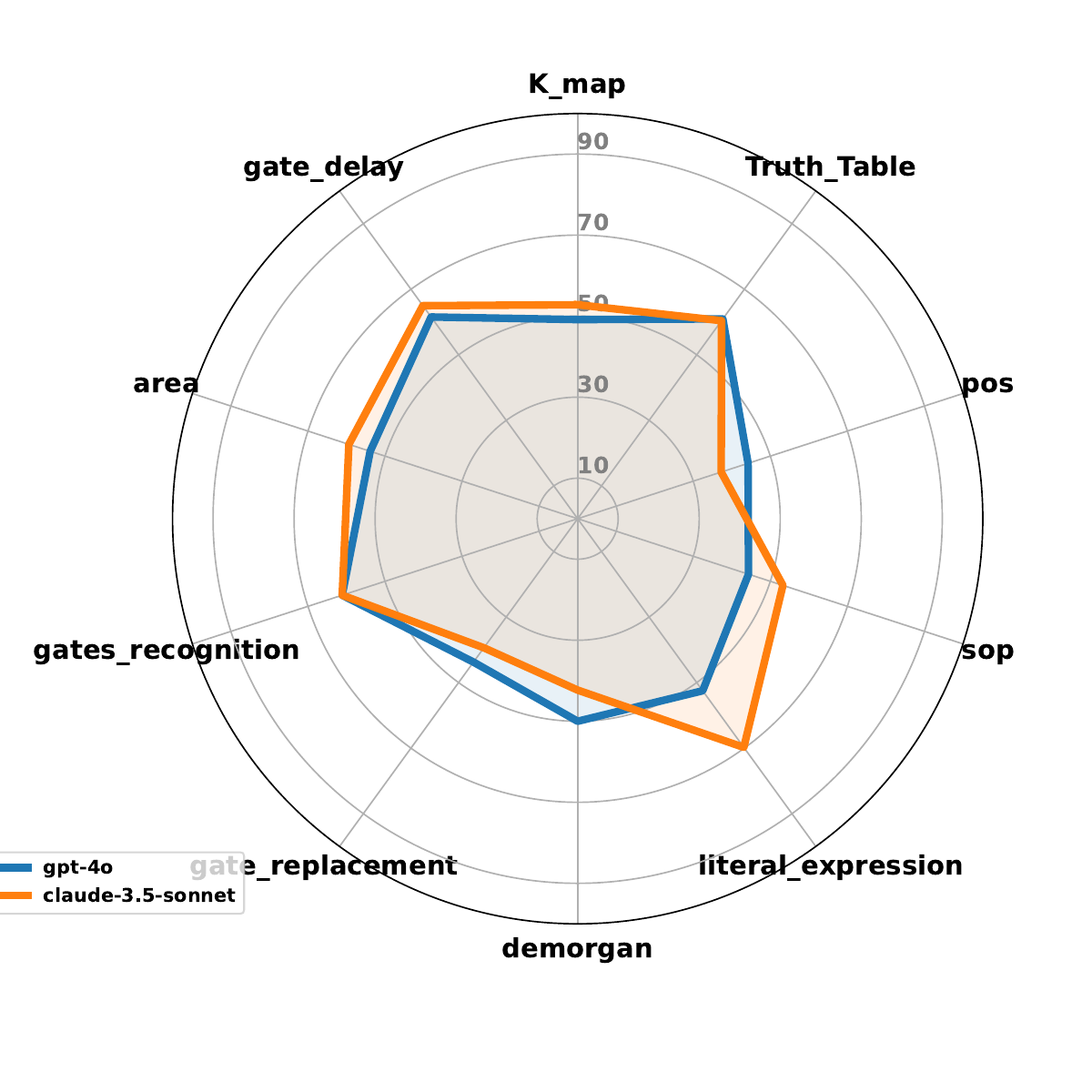}
    \subcaption{Basic concept}
    \label{leading1}
  \end{minipage}\hfil
  \begin{minipage}[t]{.32\linewidth}
    \includegraphics[width=\linewidth]{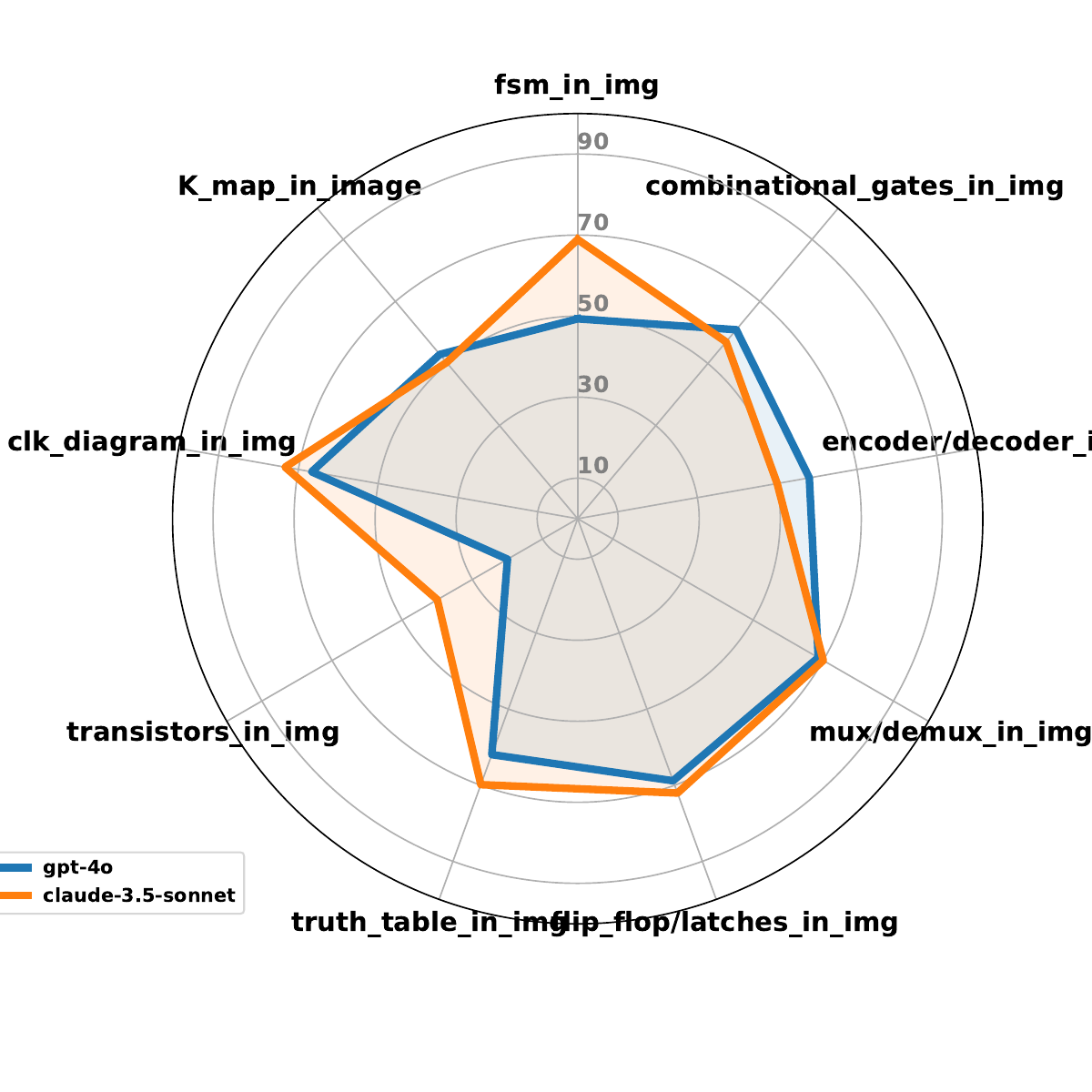} 
    \subcaption{Visual context}
    \label{leading2}
  \end{minipage}\hfil
  \begin{minipage}[t]{.32\linewidth}
    \includegraphics[width=\linewidth]{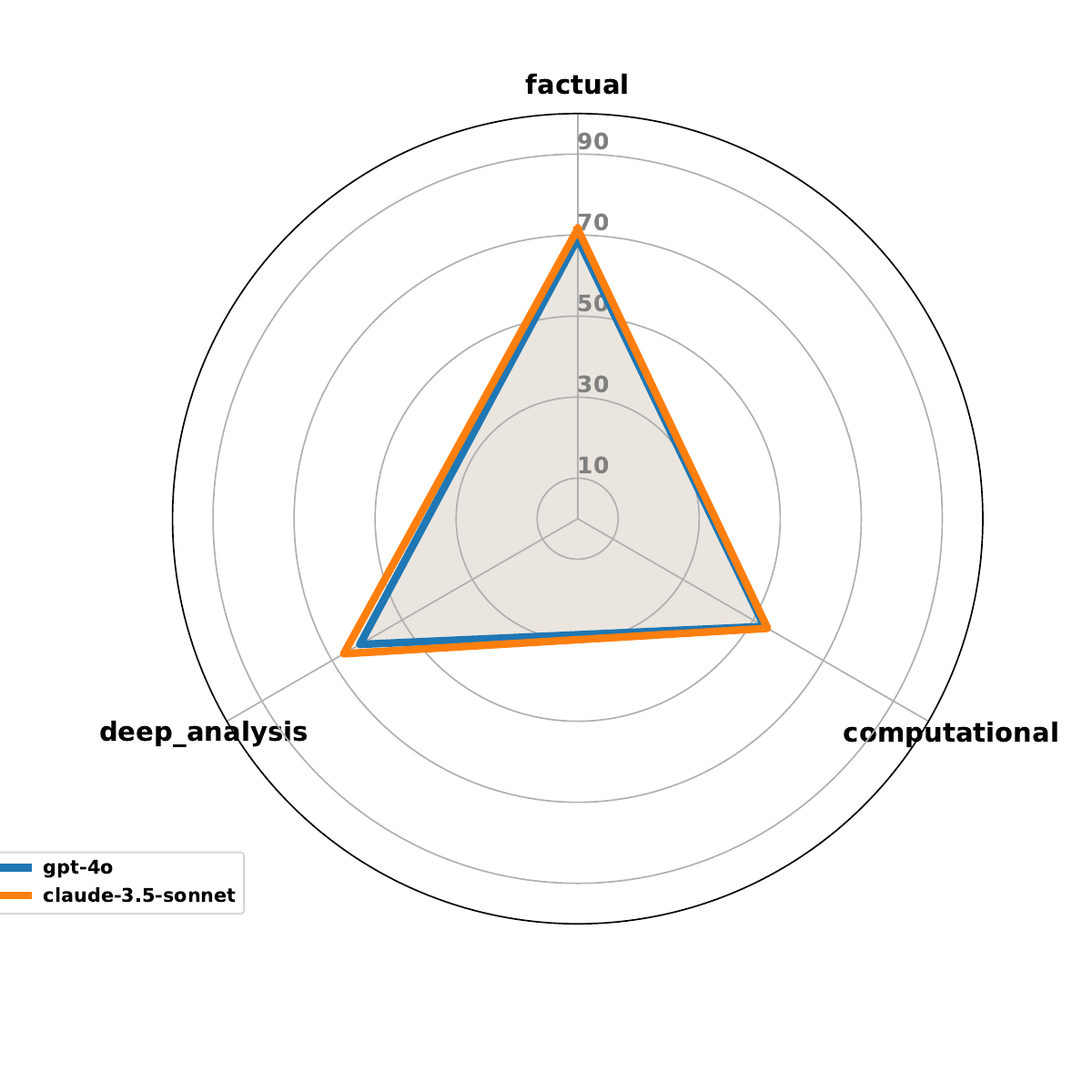}
\subcaption{Solving strategy}
\label{leading3}
  \end{minipage}%
  \caption{Performance of Multi-modal Large Language Models (MLLMs) on the Visual Question Answering (VQA) task using CoT reasoning across all primary dimensions on the complete dataset}
\end{figure*}
\begin{tcolorbox}[colback=gray!5!white,colframe=gray!75!black, colbacktitle=gray!75!black,title= Conceptual error]
        \centering
        \includegraphics[width=\textwidth]{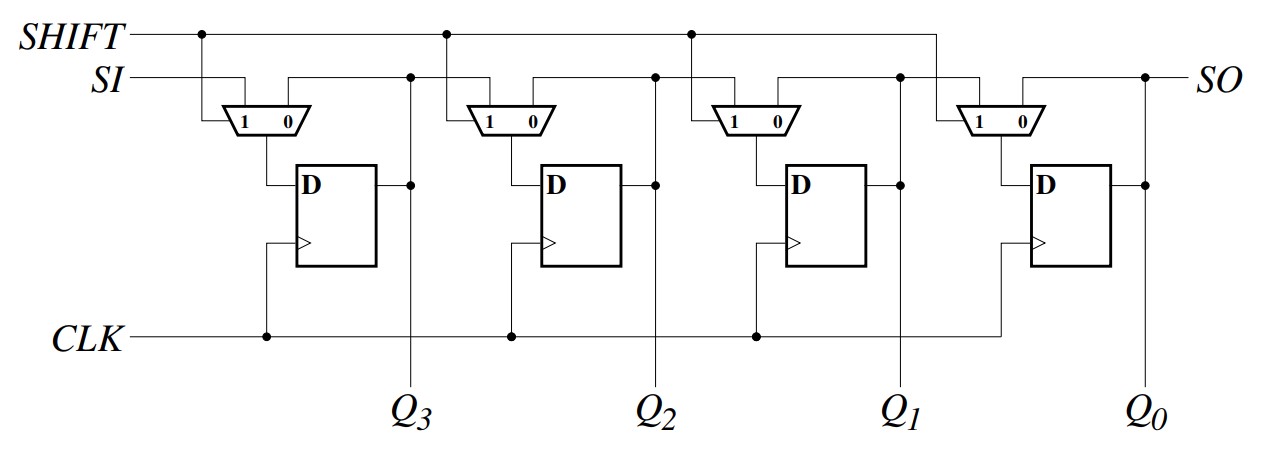} 
    \hfill
    \begin{minipage}{\textwidth} 
        \textbf{Question:} Suppose this diagram represents a set of faucets. Qi represents a water dispenser. When Qi = 1, water comes out of that faucet. If SI is set to 1 and SHIFT is low, how many clock cycles will it take for Q0 to dispense water?  \\\textbf{Choices:} (A) 1 clock cycle (B) 2 clock cycles (C) Water will continuously come out of Q0 until SHIFT is high again (D) Water will never come out of Q0 until SHIFT is high again \\
        \textbf{Correct answer:} D
    \end{minipage}
    \tcblower
    Let's analyze this step-by-step:\\ 1. The diagram shows a 4-bit shift register, where each D flip-flop represents a faucet (Q3, Q2, Q1, Q0).  2. SI (Serial Input) is the input to the first flip-flop (Q3). 3. The SHIFT line is connected to the select input of each multiplexer (1-0 box). \textbf{\textit{When SHIFT is low (0), the multiplexers select the '1' input, which is connected to the output of the previous flip-flop (or SI for the first one) (Wrong step)}} \\ 4. When SI is set to 1 and SHIFT is low:\\  - On the first clock cycle, Q3 will become 1\\   - On the second clock cycle, Q2 will become 1\\   - On the third clock cycle, Q1 will become 1\\   - On the fourth clock cycle, Q0 will become 1 \\5. Therefore, it will take 4 clock cycles for the '1' to propagate from SI to Q0.\\ Based on this analysis, the correct answer is: B) 2 clock cycles
\end{tcolorbox}

\section{Appendix A.4}\label{sec:A.4}
\subsection{Evaluation on full dataset}

\begin{table}[h!]
\centering
\begin{tabular}{c|cc}
\textbf{LLMs}      & \multicolumn{1}{c}{\textbf{\begin{tabular}[c]{@{}c@{}}V+Q\end{tabular}}} & \multicolumn{1}{c}{\textbf{\begin{tabular}[c]{@{}c@{}} E+Q\end{tabular}}} \\ \hline
\textbf{GPT-4o} & 60.39  & 72.36 \\
\textbf{Claude-3.5-sonnet }  & 62.27 & 45.21  \\

\end{tabular}
\caption{Performance of leading LLMs, GPT4-o and Claude-3.5-sonnet on complete dataset with 626 (V+Q) samples and 396 (E+Q) samples}
\label{tableleading}
\end{table}

Table \ref{tableleading} presents the performance metrics of the leading LLMs on the testmini subset. Both models exhibited accuracies around 60\% on the (V+Q) dataset, underscoring the inherent difficulty of our dataset. For the (E+Q) dataset, comprising approximately 396 samples, the results follow a similar trend as in table \ref{table1}. Figures \ref{leading1}, \ref{leading2}, and \ref{leading3} illustrate that, while the performances of Claude and GPT4-o are comparable, the Claude model generally outperforms GPT4-o across most categories in all primary problem dimensions. This indicates marginally superior adaptability of Claude to the varying demands of these dimensions in VQA tasks. Distribution for these primary problem dimensions can be found in Figure \ref{fig4}, \ref{fig5}, \ref{fig6}. 

\section{Appendix A.5}\label{sec:A.5}
Annotator compensation: The instructor that verified the dataset integrity and one of the annotators were part of the research team. The other annotator, an undergraduate research assistant, was paid \$15 per hour (well above the federally mandated minimum wage in the United States of \$7.25  per hour) for their annotation efforts.
 \end{document}